\documentclass{article}  
\usepackage{nips15submit_e,times}
 
\pdfoutput=1
\usepackage{url}
\usepackage{graphicx}
\usepackage{array}
\usepackage{amsmath}

\title{DenseBox: Unifying Landmark Localization with End to End Object Detection }

\author{
Lichao Huang$^1$ \qquad Yi Yang$^2$ \qquad Yafeng Deng$^2$ \qquad Yinan Yu$^3$ \\
$^2$Institute of Deep Learning\\
Baidu Research \\ 
$^1$\texttt{alanhuang1990@gmail.com}  \\
$^2$\texttt{\{huanglichao01,yangyi05,dengyafeng\}@baidu.com} \\
$^3$\texttt{bebekifis@gmail.com} \\
}

\nipsfinalcopy 

\begin{document}

\maketitle

%

\begin{abstract}
How can a single fully convolutional neural network (FCN) perform on object detection? We introduce DenseBox, a unified end-to-end FCN framework that directly predicts bounding boxes and object class confidences through all locations and scales of an image. Our contribution is two-fold. First, we show that a single FCN, if designed and optimized carefully, can detect multiple different objects extremely accurately and efficiently. Second, we show that when incorporating with landmark localization during multi-task learning, DenseBox further improves object detection accuray. We present experimental results on public benchmark datasets including MALF face detection and KITTI car detection, that indicate our DenseBox is the state-of-the-art system for detecting challenging objects such as faces and cars. 
 
\end{abstract}

\section{Introduction}

Our day-to-day lives are abound with instances of object detection. Checking for nearby cars during driving, finding a person, and localizing a familiar face are all examples of object detection. Object detection is one of the core problems in computer vision. Before the success of convolutional neural networks (CNNs)~\cite{krizhevsky2012imagenet}, object detection is usually addressed by sliding window based methods~\cite{felzenszwalb2010object, viola2004robust} that apply classifiers on handcrafted features~\cite{dalal2005histograms, lowe2004distinctive, cinbis2013segmentation} extracted at all possible locations and scales of images. Recently, the fully convolutional neural network (FCN)~\cite{long2015fully} based methods~\cite{sermanet2013overfeat, erhan2014scalable, YOLO} bring a revolution to the field of object detection. These FCN frameworks also follow a sliding window fashion, but their end-to-end approach of learning model parameters and image features from scratch significantly improves detection performance. 

R-CNN~\cite{girshick2014rich, girshick2015fast} further improves the accruacy on object detection beyond FCN based methods. Conceptually, R-CNN contains two phases. The first phase uses region proposal methods to generate all the potential bounding box candidates in the image. Then the second phase applies a CNN classifier to distinguish different objects for every proposal. Although R-CNN becomes the new state-of-the-art system for general object detection~\cite{everingham2010pascal, russakovsky2014imagenet}, it is very hard to detect small objects~\cite{pepik2015holding} such as human faces and far-away cars, since the low resolution and lack of contexts in each candidate box significantly decrease the classification accuracy on them. Moreover, the two different stages in the R-CNN pipeline cannot be optimized jointly, leaving the trouble for applying end-to-end training on R-CNN. 

In this work, we focus on one question: To what extent can an one-stage FCN perform on object detection? To this end, we present a novel FCN based object detector, DenseBox, that does not require proposal generation and is able to be optimized end-to-end during training. Although similar to many existing sliding window fashion FCN detection frameworks~\cite{sermanet2013overfeat, erhan2014scalable, YOLO}, DenseBox is more carefully designed to detect objects under small scales and heavy occlusion. We train DenseBox and apply careful hard negative mining techniques to boostrap the detection performance. To make it even better, we further integrate landmark localization into the system through joint multi-task learning~\cite{bengio2013representation}. To verify the usefulness of landmark localization, we manually annotate a set of keypoints for the KITTI car detection dataset~\cite{Geiger2012CVPR} and will release annotation afterward. 

Our contribution is two-fold. First, we demonstrate that a single fully convolutional neural network, if designed and optimized carefully, can detect objects under different scales with heavy occlusion extremely accurately and efficiently. Second, we show that when incorporating with landmark localization through multi-task learning, DenseBox further improves object detection accuracy. We present experimental results on public benchmark datasets including MALF (Multi-Attribute Labeled Faces) face detection~\cite{faceevaluation15} and KITTI car detection~\cite{Geiger2012CVPR}, that indicate our DenseBox is the state-of-the-art system for face detection and car detection.

\section{Related Work}

The literature on object detection is vast. Before the success of deep convolutional neural networks~\cite{krizhevsky2012imagenet}, the widely used detection systems are based on a combination of independent components. First, handcrafted image features such as HOG~\cite{dalal2005histograms,zhang2011boosted,yu2010object}, SIFT~\cite{lowe2004distinctive}, and Fisher Vector~\cite{cinbis2013segmentation} are extracted at every location and scale of an image. Second, object models such as pictorial structure model (PSM)~\cite{felzenszwalb2005pictorial} and deformable part-based model (DPM)~\cite{felzenszwalb2010object, zhu2012face, yang2013articulated} allow object parts (e.g. head, torso, arms and legs of human) to deform with geometric constraints. Finally, a classifier such as boosting methods~\cite{viola2004robust}, linear SVM~\cite{dalal2005histograms}, latent SVM~\cite{felzenszwalb2010object}, or random forests~\cite{dollar2012crosstalk} decides whether a candidate window shall be detected as containing an object.  

The application of neural networks for detection tasks such as face detection also has a long history. The first work may date back to early in 1994 when Vaillant et al.~\cite{vaillant1994original} proposed to train a convolutional neural network to detect face in image window. Later in 1996 and 1998 Rowley et al.~\cite{rowley1998neural,rowley1998rotation} presented neural network based face detection systems to detect upright frontal face in image pyramid. There is no way to compare the performance of those ancient detectors with today’s detection systems on face detection benchmarks. Even so, they are still worth revisiting, as we find many similarities in design with our DenseBox. 

Recently, several papers propose algorithms of using deep convolutional neural networks for locating objects~\cite{sermanet2013overfeat, erhan2014scalable, YOLO}. OverFeat~\cite{sermanet2013overfeat} train a convolutional layer to predict the box coordinates for multiple class-specific objects from an image pyramid. MultiBox~\cite{erhan2014scalable} generate region proposals from a network whose output layer simultaneously predicts multiple boxes, which are used for R-CNN~\cite{girshick2014rich} object detection. YOLO~\cite{YOLO} also predicts bounding boxes and class probabilities directly from full images in one evaluation. All these methods use shared computation of convolutions~\cite{sermanet2013overfeat, he2014spatial, ren2015faster, long2015fully} which has been attracting increasing attention for efficient, yet accurate, visual recognition. 

However, most state-of-the-art object detection approaches~\cite{ouyang2014deepid, li2015convolutional, erhan2014scalable,girshick2015fast,yan2015object} rely on R-CNN, which divides detection into two steps: salient object proposal generation and region proposal classification. Several recent works such as YOLO and Faster R-CNN have jointed region proposal generation with classifier in one stage or two stages. It is pointed out by~\cite{farfade2015multi} that R-CNN with general proposal methods designed for general object detection could results in inferior performance in detection task such as face detection, due to loss recall for small-sized faces and faces in complex appearance variations. They share similarities with our method, and we will discuss them with our method in more detail in later context.  

Object detection is often involved with multi-task learning such as landmark localization, pose estimation and semantic segmentation. Zhu et al.~\cite{zhu2012face} propose a tree structure model for joint face detection, pose estimation and landmark localization. Deep net based object detection systems are also natural for integrating multi-task learning. Devries et al.~\cite{devries2014multi} learn facial landmarks and expressions simultaneously through deep neural networks. Sijin et al.~\cite{li2014heterogeneous} simultaneously learn pose joint regressor and sliding window body part detector in a deep network architecture.

\section{DenseBox for Detection }
\label{sec:model} 
	\begin{figure}[!hbtp]
	\centering
	 \includegraphics[scale=0.39]{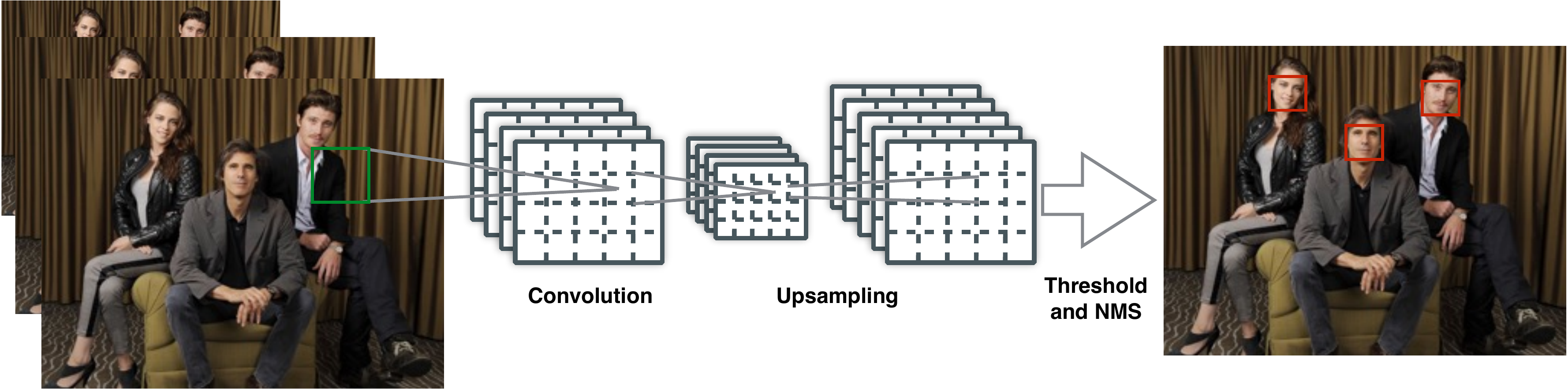}
	\caption{\textbf{The DenseBox Detection Pipeline.} 1) Image pyramid is fed to the network. 2) After several layers of convolution and pooling, upsampling feature map back and apply convolution layers to get final output. 3) Convert output feature map to bounding boxes , and apply non-maximum suppression to all bounding boxes over the threshold. }
	\label{fig:fig_overview}
	\end{figure}

The whole detection system is illustrated in Fig \ref{fig:fig_overview}. The single convolutional network simultaneously output multiple predicted bounding boxes and class confidence. All components of object detection in DenseBox are modeled as a fully convolutional network except the non-maximum suppression step, so region proposal generation is unnecessary. In the test, the system takes an image (at the size of $m\times n$) as input, and output a $ \frac{m}{4} \times \frac{n}{4} $ feature map with 5 channels.
If we define the left top and right bottom points of the target bounding box in output coordinate space as $ p_t = (x_t, y_t)$ and as $ p_b = (x_b, y_b)$ respectively, then each pixel $i$ located at $(x_i, y_i)$ in the output feature map describe a bounding box with a 5-dimensional vector $\hat{t_i } = \{ \hat{s }, \hat{dx^{t}= x_i - x_t,},\hat{dy^{t}} = y_i - y_t,\hat{dx^{b}}= x_i - x_b,\hat{dy^{b}}= y_i - y_b \}_i$ , where $\hat{s }$ is the confidence score of being an object and $\hat{dx^{t}}$, $\hat{dy^{t}}$,$\hat{dx^{b}}$, $\hat{dy^{b}}$ denote the distance between output pixel location with the boundary of target bounding box. Finally every pixel in the output map is converted to bounding box with score, and non-maximum suppression is applied to those boxes whose scores pass the threshold. 

\subsection{Ground Truth Generation} 

It is unnecessary to put the whole image into the network for training because it would take most computational time in convolving on background. A wise strategy is to crop large patches containing faces and sufficient background information for training. In this paper, we train our network on single scale, and apply it to multiple scales for evaluation. 

\begin{figure}[!hbtp]
	\centering
	 \includegraphics[scale=0.55]{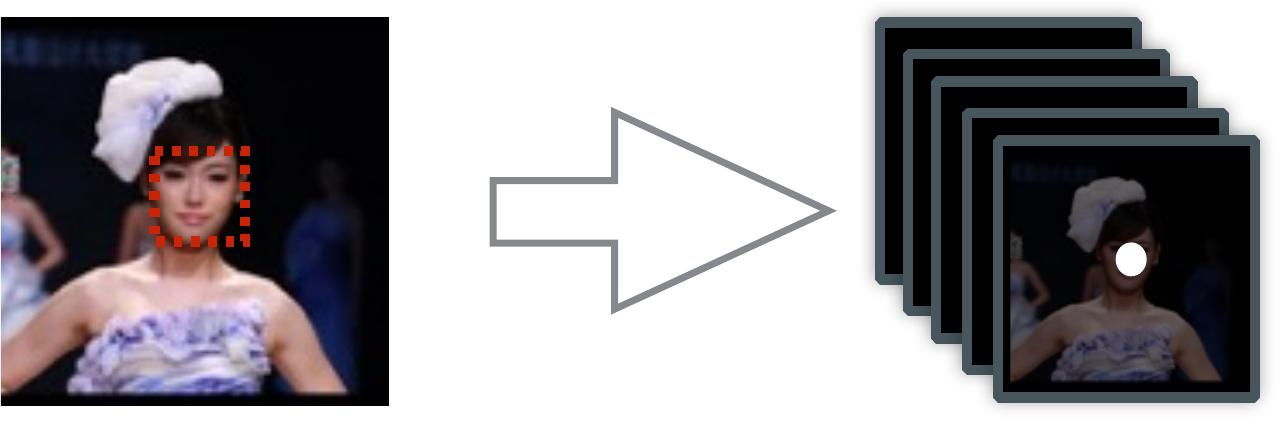}
	\caption{\textbf{The Ground Truth Map in Training .} The left image is the input patch, and the right one is its ground truth map. }
	\label{fig:fig_gt}
	\end{figure}

Generally speaking, our proposed network is trained in a segmentation-like way.   In training, the patches are cropped and resized to $240 \times 240$ with a face in the center roughly has the height of 50 pixels. The output ground truth in training is a 5-channel map sized $60 \times 60 $ , with the down-sampling factor of 4. The positive labeled region in the first channel of ground truth map is a filled circle with radius $r_c$, located in the center of a face bounding box. The radius $r_c$ is proportional to the bounding box size, and its scaling factor is set to be 0.3 to the box size in output coordinate space, as show in Fig \ref{fig:fig_gt}. The remaining 4 channels are filled with the distance between the pixel location of output map between the left top and right bottom corners of the nearest bounding box.

Note that if multiple faces occur in one patch, we keep those faces as positive if they fall in a scale range(e.g. 0.8 to 1.25 in our setting) relative to the face in patch center. Other faces are treated as negative samples. The pixels of first channel, which denote the confidence score of class, in the ground truth map are initialized with 0, and further set to 1 if within the positive label region.  We also find our ground truth generation is quite similar to the segmentation work\cite{pinheiro2015learning} by Pinheiro et.al.   In their method, the pixel label is decided by the location of object in patch, while in DenseBox, the pixel label is determined by the receptive field. Specifically, if the output pixel is labeled to 1 if it satisfies the constraint that its receptive field contains an object roughly in the center and in a given scale. Each pixel can be treated as one sample , since every 5-channel pixel describe a bounding box.

\subsection{Model Design } 

\begin{figure}[!hbtp]
	\centering
	 \includegraphics[scale=0.55]{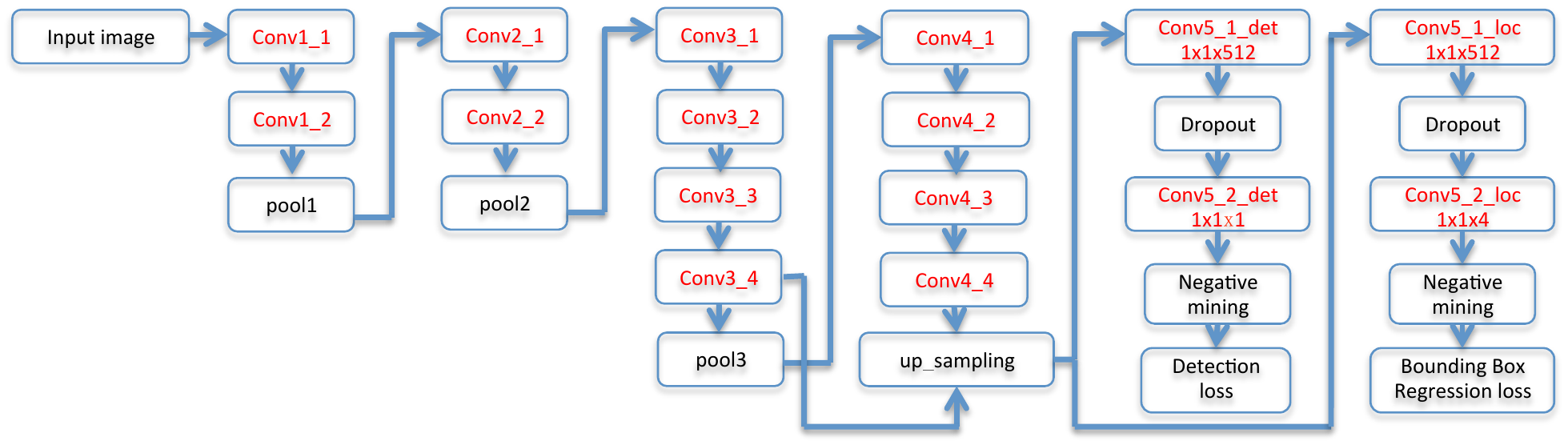}
	\caption{\textbf{Network architecture of DenseBox.} The rectangles with red names contain learnable parameters. }
	\label{fig:fig_net}
	\end{figure}

Our network architecture illustrated in Fig \ref{fig:fig_net} is derived from the VGG 19 model used for image classification\cite{simonyan2014very}. The whole network has 16 convolution layers, with the first 12 convolution layers initialized by VGG 19 model.  The output of conv4\_4 is feed into four $1 \times 1$ convolution layers, where the first two convolution layers output 1-channel map for class score, and the second two predict the relative position of bounding box by 4-channel map.  The last $1 \times 1$ convolution layers act as fully connected layers in a sliding-window fashion.

\textbf{Multi-Level Feature Fusion.} 
Recent works\cite{bertasius2014deepedge, liu2015parsenet} indicate that using features from different convolution layers can enhance performance in task such as edge detection and segmentation. Part-level feature focus on local details of object to find discriminative appearance parts, while object-level or high-level feature usually has a larger receptive field in order to recognize object. The larger receptive field also brings in context information to predict more accurate result. In our implementation, we concatenate feature map from conv3\_4 and conv4\_4. The receptive field (or sliding window size) of conv3\_4 is $48 \times 48$, almost the same size of the face size in training, and the conv4\_4 have a much larger receptive field, around $118 \times 118$ in size, which could utilize global textures and context for detection.  Note that the feature map size of conv4\_4 is half of the map generated by conv3\_4, hence we use a bilinear up-sampling layer to transform them to the same resolution. 

\subsection{Multi-Task Training.} 
\label{sec:training} 
We use the ImageNet pre-trained VGG 19 network to initialize DenseBox. Actually, in initialization, we only keep the first 12 convolution layers(from conv1\_1 to conv4\_4), and the other layers in VGG 19 are replaced by four new convolution layers with “xavier” initialization. 

Like Fast R-CNN, our network has two sibling output branches.  The first outputs the confidence score $ \hat{y}$ (per pixel in the output map) of being a target object.  Given the ground truth label $y^* \in \{0,1 \}$ , the classification loss can be defined as follows. 
	\begin{equation}\label{eq:eq_cls_loss}
	\mathcal{L} _{cls}(\hat{y},y^*) = \left \| \hat{y} - y^* \right \| ^2
	\end{equation}
Here we use $L2$ loss in both face and car detection task. We did not try other loss functions such as hinge loss or cross-entropy loss, which seems to be a more appropriate choice, as we find the simple $L2$ loss work well in our task. 

The second branch of outputs the bounding-box regression loss, denoted as $\mathcal{L} _{loc}$.  It targets on minimizing the $L2$ loss between the predicted location offsets $\hat{d} = (\hat{d}_{tx}, \hat{d}_{ty}, \hat{d}_{tx}, \hat{d}_{ty})$ and the targets  $\ d^* = (d^{*}_{tx},  d^{*}_{ty},  d^{*}_{tx}, d^{*}_{ty})$, as formulized by:
	\begin{equation}\label{eq:eq_loc_loss}
	\mathcal{L} _{loc}(\hat{d},d^*) =  \sum_{i \in \{ tx, ty,bx,by \} }  \left \| \hat{d}_{i} - d^*_{i} \right \| ^2
	\end{equation}

 \subsubsection{Balance Sampling.} 
The process of selecting negative samples is one of the crucial parts in learning. If simply using all negative samples in a mini-batch will bias prediction towards negative samples as they dominate in all samples.  In addition, the detector will degrade if we penalize loss on those samples lying in the margin of positive and negative region.  Here we use a binary mask for each output pixel to indicate whether it is selected in training. 

\textbf{Ignoring Gray Zone.} 
The gray zone is defined on the margin of positive and negative region. It should not be considered to be positive or negative, and its loss weight should be set to $0$.  For each non-positive labeled pixel in the output coordinate space, its ignore flag $f_{ign}$ is set to 1 only if there is any pixel with positive label within $r_{near} = 2$ pixel length.  

\textbf{Hard Negative Mining.} 
Analogous to hard-negative mining procedure in SVM, we make learning more efficient by searching the badly predicted samples rather than random samples. After negative mining, the badly predicted samples are very likely to be selected, so that gradient descent learning on those samples leads more robust prediction with less noise. Specifically, negative mining can be performed efficiently by online bootstrap. In the forward propagation phase, we sort the loss (Eq \ref{eq:eq_cls_loss}) of output pixels in decending order, and assign the top 1\% to be hard-negative.   In all experiments, we keep all positive labeled pixels(samples) and the ratio of positive and negative to be 1:1.  Among all negative samples, half of them are sampled from hard-negative samples, and the remaining half are selected randomly from non-hard negative. For convenience, we set a flag $f_{sel} = 1$ to those pixels (samples) selected in a mini-batch. 

\textbf{Loss with Mask.} 
Now we can define the mask $M(\hat{t}_i)$ for each sample $\hat{t}_i = \{ \hat{y}_i, \hat{d}_i \}$ as a function of flags mentioned above:
	\begin{eqnarray}\label{eq:eq_mask}
	M(\hat{t}_i) =
	\begin{cases}
	0 &  f_{ign}^{i} = 1 \text{ or } f_{sel}^{i} = 0 \\
	1 & \text{otherwise} \\
	\end{cases}
	\end{eqnarray}
 Then if we combine the classification (Eq \ref{eq:eq_cls_loss})  and bounding box regression (Eq \ref{eq:eq_loc_loss}) loss with masks, our full multi-task loss can be represented as ,
	\begin{equation}\label{eq:eq_det_loss}
	\mathcal{L} _{det}(\theta) =  \sum_{i}  \left ( M(\hat{t}_i) \mathcal{L} _{cls}(\hat{y}_i,y^*_i) + \lambda_{loc} [y^*_i >0]M(\hat{t}_i) \mathcal{L} _{loc}(\hat{d}_i,d^*_i) \right )
	\end{equation}
where $\theta$ is the set of parameters in the network, and the Iverson bracket function $[y^*_i >0]$ is activated only if the ground truth score $y^*_i$ is positive. It is obvious that the bounding box regression loss should be ignored for negative samples (background), since there is no notation for them. The balance between classification and regression tasks is controlled by the parameter $\lambda_{loc}$. In our experiments, we normalize the regression target $d^*$ by dividing by the standard object height, which is $50/4$ in ground truth map, and $\lambda_{loc} = 3$ works well in all experiments under this normalization. 

\textbf{Other Implementation Details.} 
In training, an input patch is considered to be ``positive patch" if it contains an object centered in the center at a specific scale.  These patches only contain negative samples around the positive samples.  To fully explore the negative samples in the whole dataset, we also randomly crop patches at random scale from training images, and resize them to the same size and feed them to the network. We call this kind of patch as ``random patch", and the ratio of ``positive patch" and ``random patch" in training is 1:1.  In addition, to further increase the robustness of our model, we also randomly jitter every patch before feeding them into the network.  Specifically, we apply left-right flip, translation shift (of 25 pixels), and scale deformation (from $[0.8 , 1.25]$). 

We use mini-batch SGD in training and the batch size is set to 10. The loss and output gradients must be scaled by the number of contributing pixels, so that both loss and output gradients are comparable in multi-task learning. The global learning rate starts with 0.001, and it is reduced by a factor of 10 at every 100K iterations. We follow the default momentum term weight 0.9 and the weight decay factor 0.0005.

\subsection{Refine with Landmark Localization.} 
	\begin{figure}[!hbtp]
	\centering
	 \includegraphics[scale=0.45]{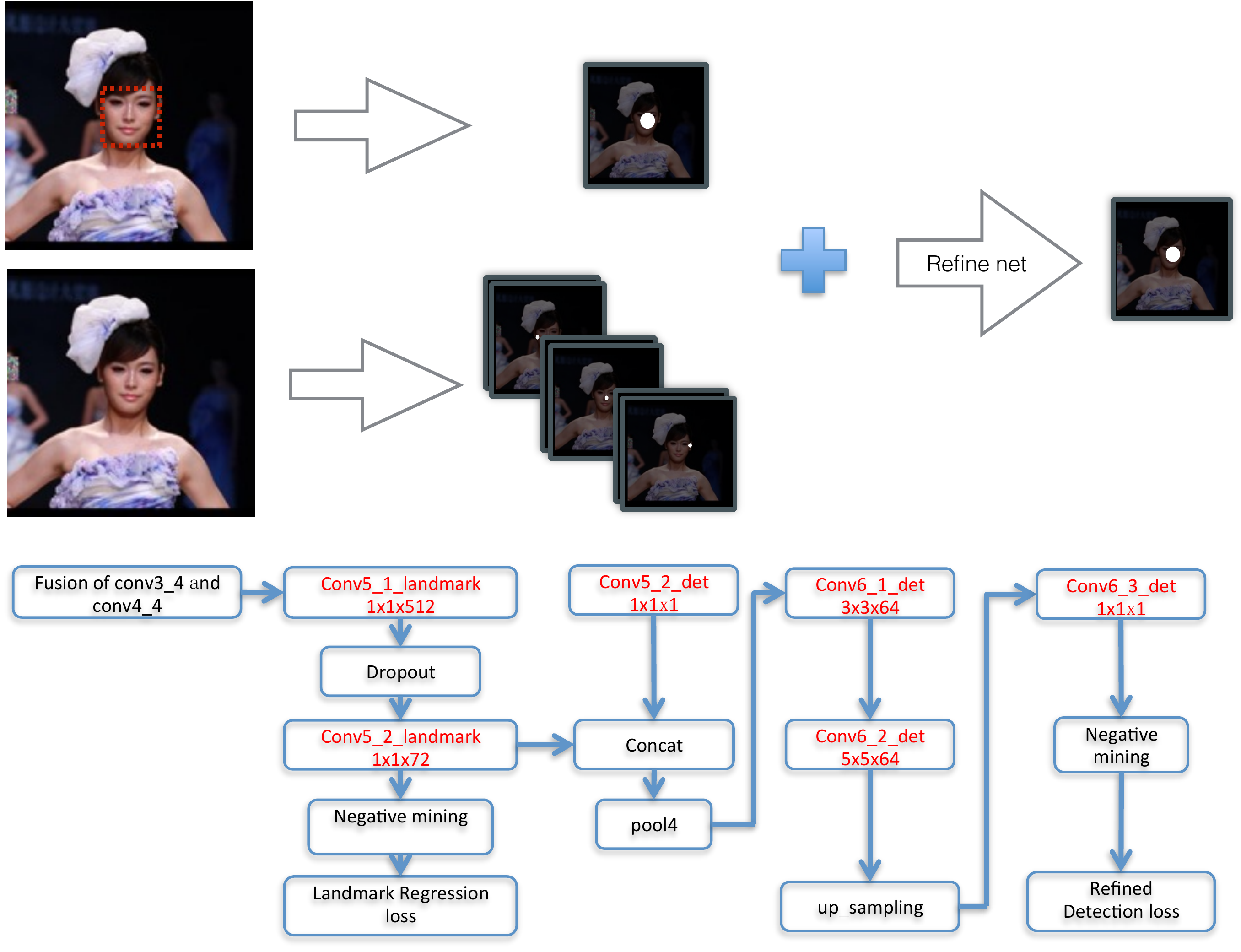}
	\caption{\textbf{Top: } The pipeline of DenseBox with landmark localization. \textbf{Bottom: } The network structure for landmark localization. }
	\label{fig:fig_refine}
	\end{figure}
In this part, we show that landmark localization can be achieved in DenseBox just by stacking a few layers owe to the fully convolution architecture. Moreover, we can refine detection results through the fusion of landmark heatmaps and face score map. As shown in Fig \ref{fig:fig_refine},  we incorporate another sibling branch output for landmark localization. Suppose there are $N$ landmarks, the landmark localization branch outputs N response maps, with each pixel represent the confidence score of being a landmark at that location.  The appearance of ground-truth maps used for this task is quite similar to the ground-truth for detection.  For a landmark instance $l^k_i$, the $i$th instance of landmark $k$, its ground-truth is a positive labeled region located at the corresponding location on the $k$th response map in the output coordinate space.  Note that the radius $r_l$ should be relative small (e.g. $r_l = 1$  )to prevent loss of accuracy.   Similar to classification task, the landmark localization loss $\mathcal{L} _{lm}$is defined as a $L2$ loss between predicted values and labels, and we still apply the negative mining and ignore region discussed in the previous section. 

The final output refine branch, taking the classification score map and landmark localization maps as input, targets on refine of the detection results.  An appropriate solution could be using high-level spatial model to learn the constraints of landmark confidence and bounding box score, to further increase the performance of detections. Tompson et.al.\cite{tompson2014joint} proposed a MRF-like model using modified convolution (SoftPlus convolution) with non-negative output to connect the distribution of spatial location for each body part.  However, their model also include $Log$ and $Exp$ stages, which make model difficult to train.  In our implementation, we use convolutions with ReLU activation to approximate the spatial model.  If we denote the refine detection loss as $\mathcal{L} _{rf}$, which is almost the same as the classification loss $\mathcal{L} _{cls}$ mentioned before but the predict map is from the refine branch, the full loss becomes as ,
	\begin{equation}\label{eq:eq_full_loss}
	\mathcal{L} _{full}(\theta) = \lambda_{det} \mathcal{L} _{det}(\theta) + \lambda_{lm} \mathcal{L} _{lm}(\theta) + \mathcal{L} _{rf}(\theta)
	\end{equation}
where $\lambda_{det}$ and $\lambda_{lm}$ controll the balance of the three tasks. They are assigned to 1 and 0.5 respectively in our experiments. 

\subsection{Comparison } 
The highlight of DenseBox is that it frames object detection as a regression problem and provides an end-to-end detection framework.  Several recent works such as YOLO and Faster R-CNN have jointed region proposal generation with classifier together. Here we compare DenseBox to other related detection systems, pointing out the key similarities and differences. 
 
\textbf{Traditional NN-based Face Detector.} The neural network-based face detectors refer to those face detection system using neural network before the recent break-through results of CNNs for image classification. Applying neural networks for face detection has a long history, and the early works date back to 1990s\cite{vaillant1994original}.  Rowley et al.\cite{rowley1998neural} train neural network-based detectors which only is activated on faces with specific size, and apply detectors on the image pyramid with sliding-window fashion. Our DenseBox is very similar to them in the detection pipeline, excepting that we use modern CNNs as detectors. Hence the DenseBox could be called as `` Modern NN-based detector“ in one sense. 

\textbf{OverFeat.} OverFeat\cite{sermanet2013overfeat} designed by Sermanet et al. might be the first work that train a convolution neural network to perform classification and localization together after the success application of deep CNNs for image classification\cite{krizhevsky2012imagenet}. It also apply fully convolutional network on test time, an equivalent but much efficient way to perform sliding window detection. However it still disjoints classification and localization in training, and need complex post-processing to produce detection results. Our method is very similar to OverFeat but a multi-task jointly learned end-to-end detection network. 

\textbf{Deep Dense Face Detector (DDFD)} The DDFD, psoposed by Farfade et.al.\cite{farfade2015multi},  is a face detection system based on convolutional neural networks. It claims to have superior performance over R-CNN on face detection task due to the reason that proposal generation in R-CNN may miss some face regions.  Although the DDFD is a complete detection pipeline,  the DDFD is not an end-to-end framework since it separate the class probability prediction and bounding box localization as two tasks and two stages.  Our DenseBox can be optimized directly for detection , and can be easily improved by incorporating landmark information. 

\textbf{Faster R-CNN.} The faster R-CNN\cite{ren2015faster} still use region proposals to find objects in an image. Unlike the its former variants, the region proposals in faster R-CNN is produced by region proposal networks(RPNs) sharing convolutional feature computation with classifiers in the second stage. The PRN shares many similarities with our method DenseBox. However, The PRN needs predefined anchors while ours does not. The PRN is trained on multi-scale objects while the DenseBox presented in this paper is trained on one scale with jitter-augmentation, which means our method need to evaluate feature at multiple scales. Moreover, the training schemes are quite different between DenseBox and PRN. 

\textbf{MultiBox.} The MultiBox\cite{erhan2014scalable} trains a convolutional neural network to generate proposals instead of selective search. Both DenseBox and MultiBox are trained to predict bounding boxes in an image, they generate bounding boxes in different way.  As compared in \cite{ren2015faster} , the MultiBox method generates 800 non-translation-invariant anchors, whereas our DenseBox output translation-invariant bounding boxes like RPN. As the down-sampling factor of output map is 4,  DenseBox will densely generate one bounding box with score at every 4 pixels. 

\textbf{YOLO.}  Redmon et al.\cite{YOLO} propose a unified object detection pipeline, called YOLO. Both DenseBox and YOLO can be trained end-to-end from images, but the model design differs in the output layers.  The YOLO system takes a $448 \times 448$ image as input, and outputs $7 \times 7$ grid cells, only 49 bounding boxes per image. Our DenseBox uses up-sampling layers to keep a relative high-resolution output, with a down-sampling scale factor of 4 in our model. This enables our network capable to detect very small objects and highly overlapped objects, which YOLO is unable to deal with.

\section{Experiments}

In this section, we demonstrate the performance of DenseBox on MALF(Multi-Attribute Labelled Faces) dataset\cite{faceevaluation15}  and KITTI\cite{Geiger2012CVPR} car detection task. We also evaluate our method on those tasks with or without the help of landmark annotation, showing that multi-task learning with landmark localization can significantly boost the performance. We compare our results with current the state-of-the-art systems, which shows that our method achieves competitive results on object detection tasks.  Nothe that we do not compare the performances of our DenseBox with original R-CNN directly on those task, but we highlight the performances of other methods which claim to use R-CNN or those methods have alrealy compared themselves to R-CNN. 

\subsection{ MALF  Detection Task}
The MALF detection test dataset contains 5,000 images in collected from the Internet. Unlike the widely used FDDB\cite{jain2010fddb} face detection benchmark, which is collected from news photos and the pose tends to be frontal, the face images in MALF have much larger diversity, making it closer to real world application than FDDB. 

\begin{figure} 
	\centering
	 \includegraphics[scale=0.45]{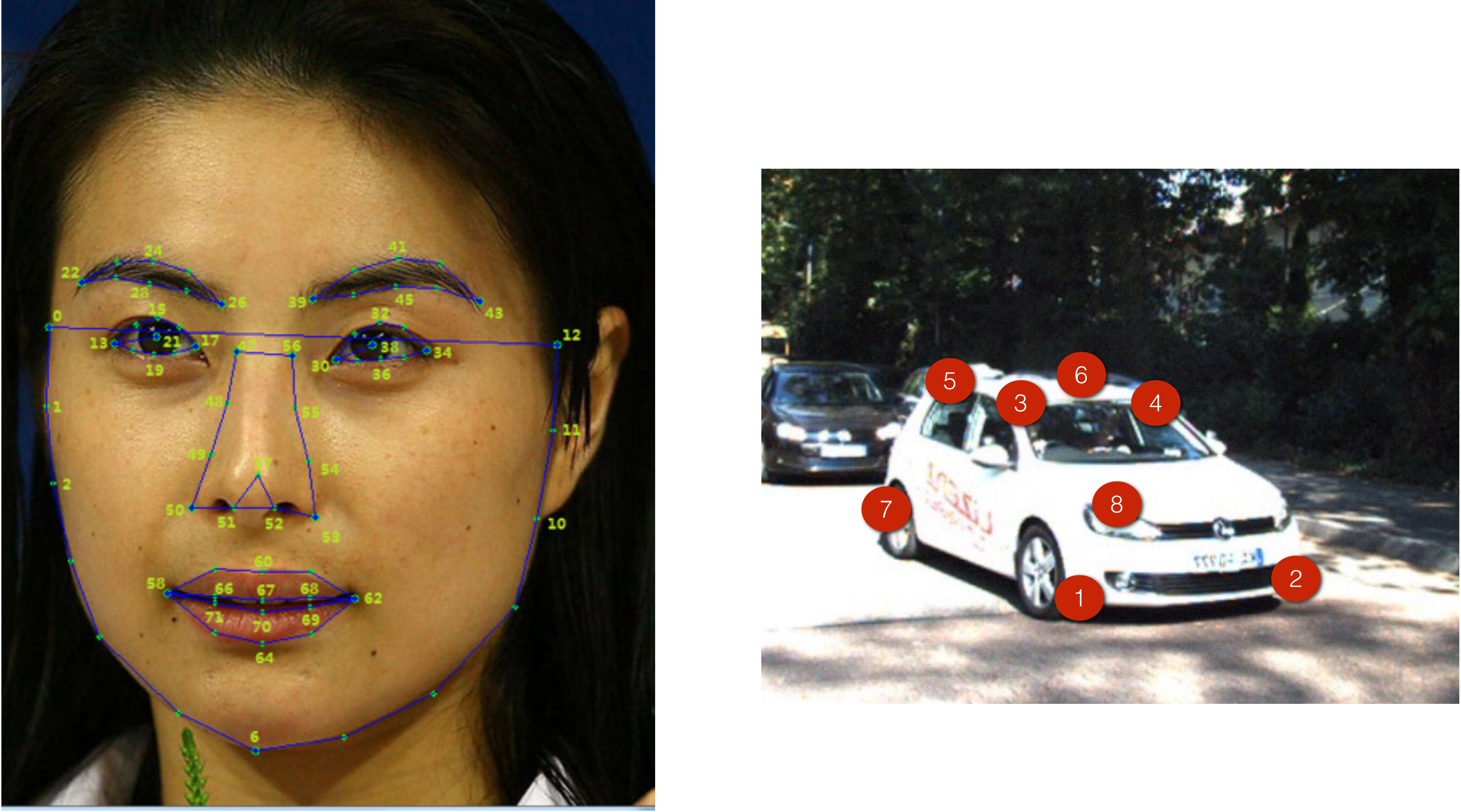}
	\caption{\textbf{Left: } 72 landmarks for face. \textbf{Right: } 8 landmarks for car. }
	\label{fig:fig_landmark}
	\end{figure}

\begin{figure} 
  \begin{minipage}[t]{0.5\linewidth} 
    \centering 
    \includegraphics[scale = 0.5]{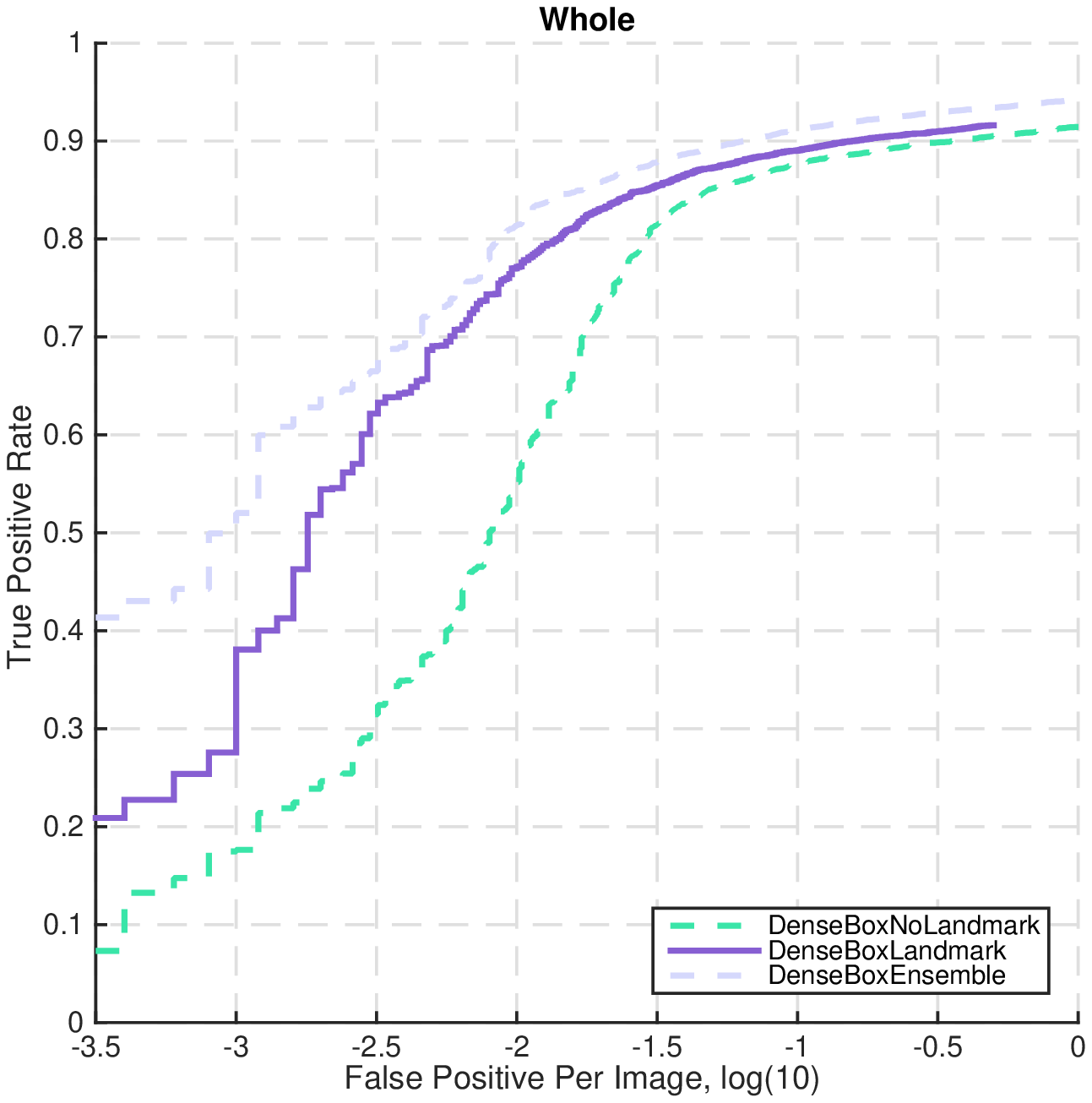}  (a)
  \end{minipage}%
  \begin{minipage}[t]{0.5\linewidth} 
    \centering 
    \includegraphics[scale = 0.5]{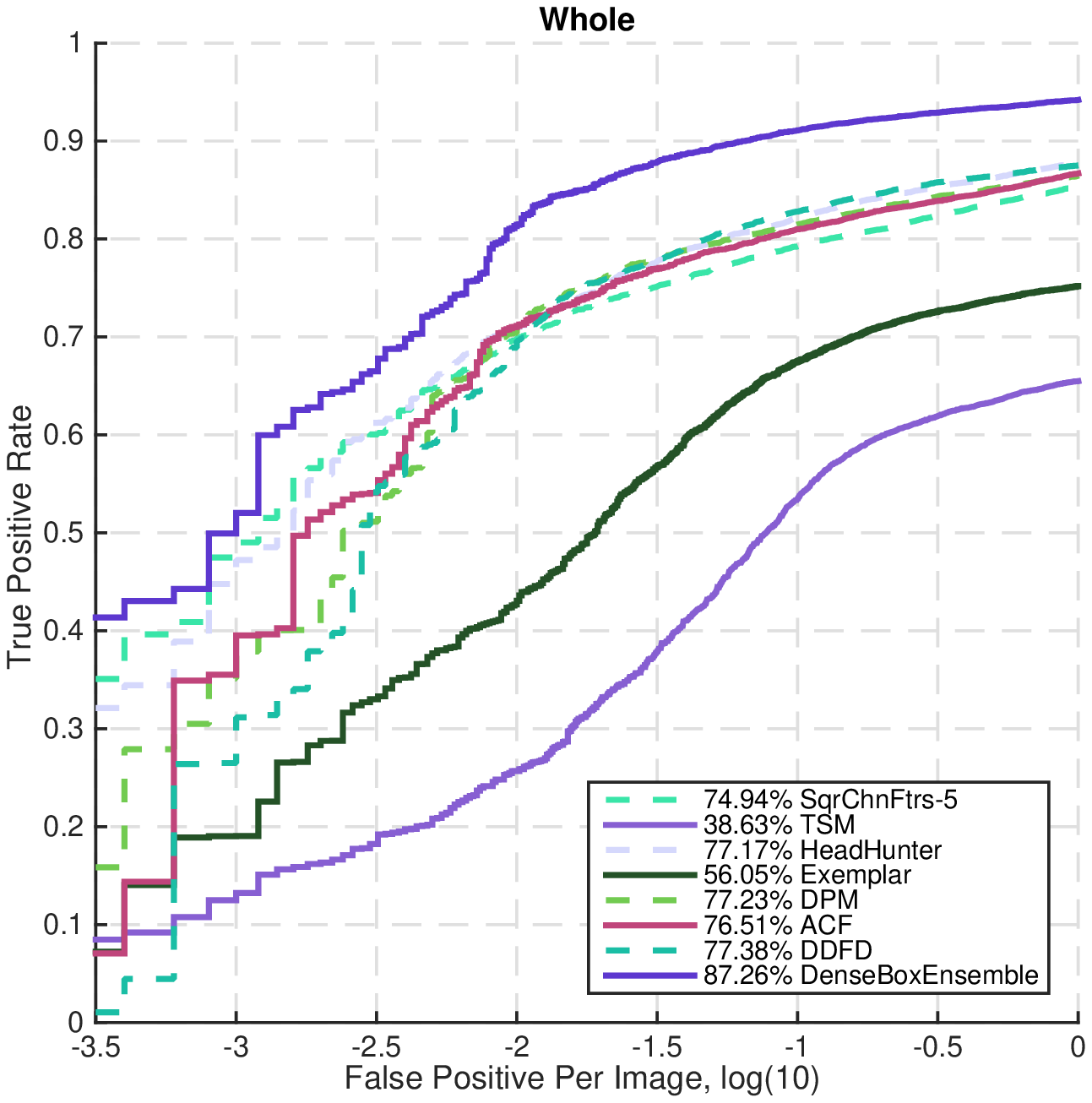}  (b)
  \end{minipage} 
 \caption{  \textbf{Result on MALF dataset. }  (a) Comparison of different versions of DenseBox; (b)The curves and mean recall rate of DenseBox and other methods;  }\label{fig:fig_malf}
\end{figure}

\textbf{Training and Testing.} We train two models described in section \ref{sec:model} on 31,337 Internet-collected  images with  81,024 faces annotated with 72 landmarks illustrated in Fig \ref{fig:fig_landmark}. One model only use bounding box information, while the other model utilize both bounding box and landmark information for comparison. They are both initialized with ImageNet pre-trained VGG19 model.  The faces in training are roughly scaled to 50 pixels in height, and the scale jitter range is $[0.8,1.25]$ , the same as described in section \ref{sec:training}.  On testing, we first selectively down sample images so that for each image the longest image side does not exceed 800 pixels.  Then we test our model on each image at several scales.  The test scale starts from $2^{-3}$ to $2^{1.2}$ with the step of $2^{0.3}$. This setting enable our models to detect faces from 20 pixels to 400 pixels in height. The non-maximum suppression IOU threshold in face detection is set to $0.5$.  Under this configuration, it taks several seconds to process one image in MALF dataset on an Nvidia K40 GPU. 
\textbf{Results.} 
We illustrate the results of three versions of DenseBox on MALF dataset. The ``DenseBoxNoLandmark’’ denotes DenseBox without landmark in training. ``DenseBoxLandmark’’ is the model incorporating landmark localization, and ``DenseBoxEnsemble’’ is the result of ensembling 10 DenseBox with landmarks from different batch iterations. As shown in Fig \ref{fig:fig_malf}, landmark localization gives a significant performance boost on face detection.  We also notice that the models trained with different batch iterations still have high diversity since another significant boost has been seen by model ensemble. Then we compare our best model with other state-of-the-art methods on MALF. Surprisingly, our model achieves the best performance, with mean recall rate of $87.26\%$, almost outperform DDFD by $10\%$, which claims to have better performance than R-CNN on face detection task.

\subsection{ KITTI Car Detection Task}
The KITTI object detection benchmark consists of 7481 training images and 7518 test images. The total number of objects in training sums up to 51,867, in which cars only accounts for 28,742. The key difficulty of KITTI car detection task is that a great amount of cars are in small size (height < 40 pixels) and occluded.  To overcome this difficulty,  previous works such as\cite{li2014integrating} need careful part and occlusion modeling. 

\textbf{Training and Testing.} As well as in face detection task, we train two models(one without landmark, the other with landmark) on the KITTI object detection training set. Since KITTI does not provide landmarks for car, we selectively annotate 8 landmarks for large cars (height > 50 pixels). The landmarks of car is shown in Fig \ref{fig:fig_landmark}, and we finally annotate 7790 cars, roughly $27\%$ of the total cars. The testing procedure is the same as in face detection, except that we do not down sample car images. The evaluation metric of KITTI car detection task is different from general objecte detection. KITTI requires an overlap of 70\% for true positive bounding box, while other tasks such as face detection only requires 50\% overlap. This strict criteria requests high accurate car localization. On KITTI, we set the non-maximum suppression  IOU threshold to 0.75. 

\textbf{Results.} 

	\renewcommand{\arraystretch}{1.5}
	\begin{table}[!hbt]
	\centering
	\begin{tabular}{|l|c|c|c|}
	\hline

	Method & \textit{Moderate} &\textit{Easy} & \textit{Hard}\\
	\hline
	Regionlets~\cite{long2015accurate}	 &76.45\%	  &84.75\%	  &59.70\% 	 \\ 
	AOG~\cite{li2014integrating}	&74.26\%	&84.24\% 		&60.51\% \\
	3DVP~\cite{xiang2015data}			&75.77\%	&87.46\%	&65.38\%       \\
	spCov\_LBP 	&77.40\%	&87.19\%	&60.60\%   \\
	DeepInsight	&84.40\%	&84.59\%	&76.09\%     \\
	NIPS ID 331	&87.14\%	&88.33\%	&76.11\% \\
	DJML	&88.79\%	&91.31\%	&77.73\% \\
	\hline
	
	\hline
	DenseBox (without landmark)	&85.07\%	&82.33\%	&76.27\%   \\
	DenseBox (withlandmark)		&85.74\%	&83.63\%	&76.71\%     \\
 
	\hline
	\end{tabular}
	\caption{The Averate Presision on KITTI Car Detection Task} 
	\label{tab:tab_kitti}
	\end{table}

\textbf{Results.} Table \ref{tab:tab_kitti} shows the results of DenseBox and other methods. We can see that partially annotated landmark information (27\%) still can boost detection performance. On average, model with landmark localization slightly outperforms no-landmark model by 0.9\% in average precision.  The promotion on performance is not as great as in face detection.  The reason could be that the landmark information is not sufficient.  The insufficient lies on both the amount (27\% in car while 100\% in face) and the quality (8 landmarks in car whereas 74 in face). Compared with other methods, the DenseBox still achieves competitive results.  DenseBox defeats traditional detection system such as Regionlets and spCov by a large margin. Our average precision on moderate car is 85.74\%, slightly better than DeepInsight, which use R-CNN framework with ImageNet pre-trained GoogLeNet\cite{szegedy2014going}.  Our model has been ranked as the top 1 for 4 months until an anonymous submission titled  ``NIPS ID 331’’, which use stereo information for training and testing. Recently a method named ``DJML’’ overtakes all other methods. 

	\begin{figure}[!hbtp]
	\centering
	 \includegraphics[scale=0.4]{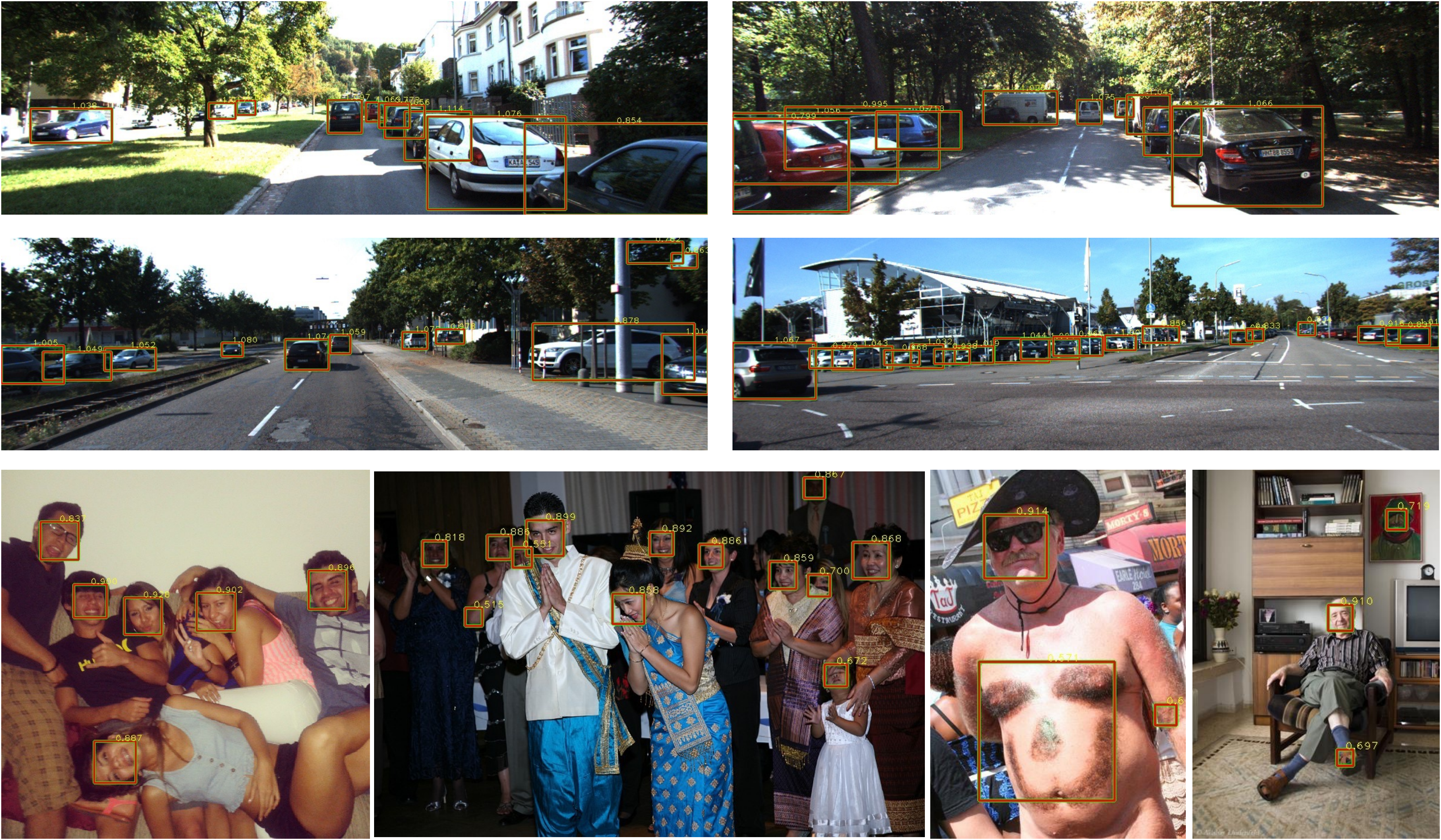}
	\caption{\textbf{Examples on both the MALF detection set KITTI car detection set. }The numbers above the bounding boxes are the confidence score. Our system works very well in complex scene where objects are small and highly occluded. However, it still could miss some objects and generate false alarm.     }
	\label{fig:fig_case}
	\end{figure}
\section{Conclusion}

We have presented the DenseBox, a unified end-to-end detection pipeline for detection. The performance can be boosted easily by incorporating landmark information. We also analysis our method and other related object detection system, highlighting the difference and the contribution of DenseBox.  The DenseBox achieves impressive performance on both face detection and car detection task, demonstrating its high suitable for situation where proposal generation might fail. The key problem of DenseBox is the speed.  The original DenseBox presented in this paper needs several seconds to process one image. But this has been addressed in our latter version.  We will present another paper describing a real-time detection system on KITTI and face detection, called DenseBox2.

 \bibliographystyle{abbrv}
 \bibliography{references.bib}
\end{document}